\pdfoutput=1
\documentclass{article}
\usepackage[preprint]{neurips_2020}
\usepackage{epstopdf}
\usepackage{times}
\usepackage[utf8]{inputenc} 
\usepackage[T1]{fontenc}    
\usepackage{hyperref}       
\usepackage{url}            
\usepackage{grffile}
\usepackage{booktabs}       
\usepackage{amsfonts,amsmath,amssymb}       
\usepackage{nicefrac}       
\usepackage{microtype}      
\usepackage{multirow}
\usepackage{subfigure}
\usepackage{algorithm,algorithmic}
\usepackage{xcolor}
\usepackage[normalem]{ulem}
\usepackage{arydshln}
\usepackage{caption}
\usepackage{wrapfig}
\usepackage{graphicx}
\usepackage{kvsetkeys}

\newcommand{\modeltitle}{Set-based Task-Adaptive Meta-Pruning}
\newcommand{\fulltitle}{Rapid Structural Pruning of Neural Networks with \modeltitle}
\newcommand{\shorttitle}{STAMP}
\title{\fulltitle}

\newcommand{\bsy}{\boldsymbol}

\begin{document}
\author{ Minyoung Song$^{1}$, Jaehong Yoon$^{1}$, Eunho Yang$^{1,2}$, Sung Ju Hwang$^{1,2}$\\
KAIST$^{1}$, AITRICS$^{2}$, South Korea\\
 \texttt{\{mysong105, jaehong.yoon, eunhoy, sjhwang82\}@kaist.ac.kr}}

\maketitle

\begin{abstract}
As deep neural networks are growing in size and being increasingly deployed to more resource-limited devices, there has been a recent surge of interest in network pruning methods, which aim to remove less important weights or activations of a given network. A common limitation of most existing pruning techniques, is that they require pre-training of the network at least once before pruning, and thus we can benefit from reduction in memory and computation only at the inference time. However, reducing the \emph{training} cost of neural networks with rapid structural pruning may be beneficial either to minimize monetary cost with cloud computing or to enable on-device learning on a resource-limited device. Recently introduced random-weight pruning approaches can eliminate the needs of pretraining, but they often obtain suboptimal performance over conventional pruning techniques and also does not allow for faster training since they perform unstructured pruning. To overcome their limitations, we propose \emph{Set-based Task-Adaptive Meta Pruning (STAMP)}, which task-adaptively prunes a network pretrained on a large reference dataset by generating a pruning mask on it as a function of the target dataset. To ensure maximum performance improvements on the target task, we \emph{meta-learn} the mask generator over different subsets of the reference dataset, such that it can generalize well to any unseen datasets within a few gradient steps of training. We validate STAMP against recent advanced pruning methods on benchmark datasets, on which it not only obtains significantly improved compression rates over the baselines at similar accuracy, but also orders of magnitude faster training speed.
\end{abstract}

\section{Introduction}
Deep learning has achieved remarkable progress over the last years on a variety of tasks, such as image classification~\cite{krizhevsky2012imagenet, wang2017residual, rawat2017deep}, object detection~\cite{lin2017feature, liu2020deep}, and semantic segmentation~\cite{lin2017refinenet,huang2019ccnet}. A key factor to the success of deep neural networks is their expressive power, which allows them to represent complex functions with high precision. Yet, such expressive power came at the cost of increased memory and computational requirement. Moreover, there is an increasing demand to deploy deep neural networks to resource-limited devices, which may not have sufficient memory and computing power to run the modern deep neural networks. Thus, many approaches have been proposed to reduce the size of the deep neural networks, such as network pruning, training the model with sparsity-inducing regularizations or prior~\cite{han2015learning,yoon2017combined,lee2018adaptive}, network distillation~\cite{HintonG2014nipsw,hui2018fast}, and network quantization~\cite{han2015deep,jung2019learning}. Arguably the most popular approach among them is network pruning, which aims to find the optimal subnetwork that is significantly smaller than the original network either by removing its weights and activations (\emph{unstructured}) or filters and channels (\emph{structured}). Structured pruning is often favored over unstructured pruning since GPUs can exploit its data locality to yield actual reduction of inference time, while unstructured pruning sometimes lead to longer inference time than the full networks~\cite{wen2016learning}.  

Yet, most conventional pruning techniques have a common limitation, in that they require a network pretrained on the target dataset. With such two-stage schemes, training will inevitably take more time than training of a full network, and thus most works focus only on the efficiency at inference time. However, in many real-world scenarios, it may be desirable to obtain \emph{training-time speedups} with pruning. For instance, if we have to train a large network for a large dataset on cloud, it may incur large monetary cost (Figure~\ref{fig:con}(a)). As another example, due to concerns on data privacy, we may need to train the network on resource-limited devices (Figure~\ref{fig:con}(b)), but the device may not have enough capacity even to load the original unpruned networks on memory. Handling such diverse requirement efficiently for each end user is crucial for a success of a machine learning platform (Figure~\ref{fig:con}). Then how can we perform pruning without pretraining on the target task? 

A few recently introduced methods, such as SNIP~\cite{lee2018snip} and Edge-Popup~\cite{ramanujan2019s} allow to prune randomly initialized neural networks, such that after fine-tuning, the pruned network obtains performance that is only marginally worse than that of the full network. This effectively eliminates the needs of pretraining, and SNIP further reduces pruning time by performing pruning in a single forward pass. However, they are limited in that they perform \emph{unstructured pruning} which will not result in meaningful speedups on GPUs, either at inference or training time. Moreover, they underperform state-of-the-art structure pruning techniques with pretraining. Thus, none of the existing works can obtain strucutrally pruned subnetworks that provide us practical speedups both at the training and inference time, with minimal accuracy loss over the full network. 

To achieve this challenging goal, we first focus on that in real-world scenarios, we may have a network pretrained on a large reference dataset (Figure~\ref{fig:con} (c)). If we could prune such a reference pretrained network to make it obtain good performance on an \emph{unseen} target task, it would be highly efficient since we only need to train the model once and use it for any given tasks. However, pruning a network trained on a different dataset may yield a suboptimal subnetwork for the target task. Thus, to ensure that the pruned network obtains near-optimal subnetwork for an unseen task, we propose to \emph{meta-learn} the task-adaptive pruning mask generator as a \emph{set function}, such that given few samples of the target dataset, it instantly generates a task-optimal subnetwork of a pretrained reference network.

We validate our \emph{Set-based Task-Adaptive Meta Pruning (STAMP)} on multiple benchmark datasets against recently proposed structural pruning and random-weight pruning baselines. The results show that our method can rapidly prune an network to obtain a network is significantly more compact than the networks with similar accuracy using baseline pruning techniques. Further, this rapid structural pruning allows our model to significantly reduce the training cost in terms of both memory, computation, and wall-clock time, with minimal accuracy loss. Such efficiency makes STAMP appealing as a cheap alternative for neural architecture search in machine learning platforms (See Figure~\ref{fig:con}). The contribution of our work is threefold:


\begin{figure*}
\small
    \centering
    \begin{tabular}{c }\hspace{-0.1in}

       \includegraphics[width=10cm]{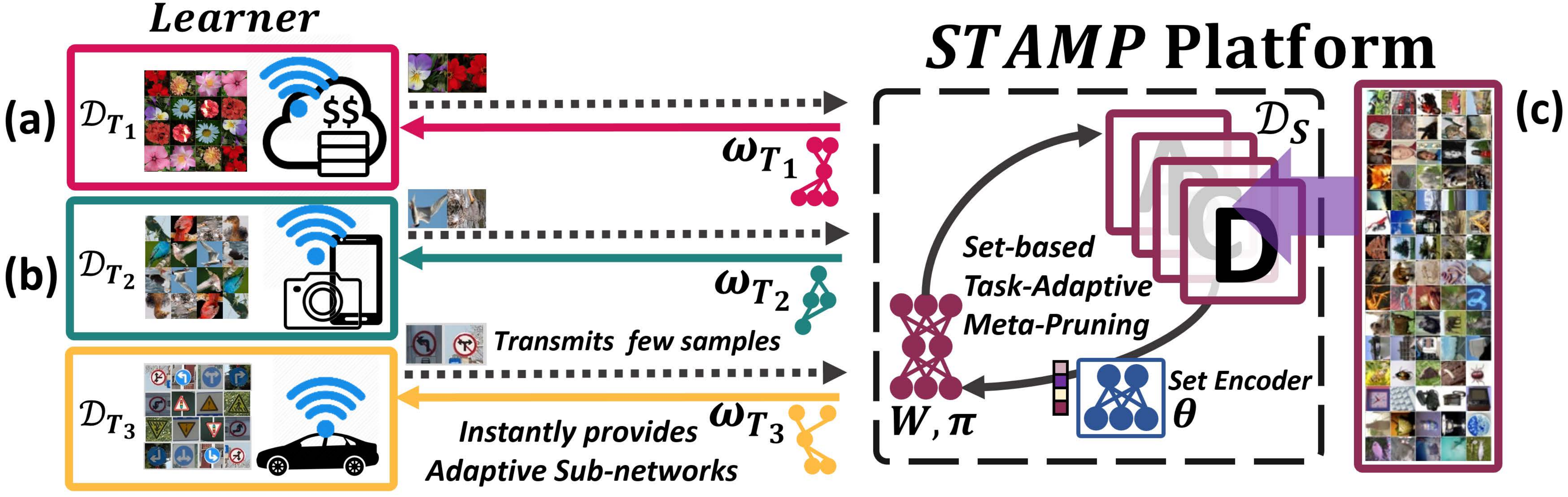}\\

    \end{tabular}
    \caption{\small \textbf{Illustrations of our Set-based Task-Adaptive Meta-Pruning (STAMP):} \emph{STAMP} meta-learns a general strategy to rapidly perform structural pruning of a reference network, for unseen tasks. If a learner gives a small fraction of information for his/her target tasks, STAMP almost instantly provides an optimally pruned network architecture which will train faster than the full network with minimal accuracy loss.}    
    \label{fig:con}
    \vspace{-0.1in}
\end{figure*}

\begin{itemize}
 \item We propose a novel set-based structured pruning model, which instantly generates a pruning mask for a given dataset to prune a target network.
 
 \item We suggest a meta-learning framework to train our set-based pruning mask generator, to obtain an approximately optimal subnetwork within few gradient steps on an unseen task. 
 
 \item We validate our meta-pruning model on benchmark datasets against structured and random weight pruning baselines, and show that it obtains significantly more compact subnetworks, that require only a fraction of wall-clock time to train the network to target accuracy.
\end{itemize}

\section{Related Work}
\paragraph{Neural network pruning.} During recent decades, there has been a surge of interest on weight pruning schemes for deep neural networks to promote memory/computationally efficient models. 
Unstructured pruning prunes the weight of the network without consideration of its structure. Some unstructured pruning methods have been shown to obtain extremely sparse networks that match the accuracy of full network, such as iterative magnitude pruning~\cite{han2015learning} which repeats between training and finetuning to recover from the damage from pruning. 
Lottery Ticket Hypothesis (LTH)~\cite{frankle2018lottery} discusses the existence of a subnetwork which matches the accuracy and training time of a full network, referred as the \textit{winning ticket}, and show that they can be found with iterative magnitude pruning. 
SNIP~\cite{lee2018snip} propose a simple pruning method which can identify a similar subnetwork without pretraining in single forward step. Though SNIP does not strictly find a \textit{winning ticket}, it is highly efficient and opens possibility to further research on rapid pruning without pretraining. Edge-Popup~\cite{ramanujan2019s} finds optimal subsets from random weights, without any pretraining, which is also simple. However, SNIP is faster than Edge-popup in searching a pruned network.
 
Although unstructured pruning methods find extremely sparse subnetworks and gets simpler, due to poor data locality, it is difficult to reduce the network inference time on general-purpose hardware. Due to this limitation, recent works \cite{liu2017learning,liu2019metapruning,he2017channel,guo2020channel,luo2017thinet,zhuang2018discrimination} target to prune groups of weights (e.g., channels or neurons), to achieve actual reduction in the model size. Such structured pruning methods are useful in a resource-limited environment with compressed architectures to practically reduce the memory requirement and the running time and at inference time. SSL~\cite{wen2016learning} introduces a structured sparsity regularization method to prune neurons using (2,1)-norm during training. CGES~\cite{yoon2017combined} propose to combine group sparsity with exclusive sparsity regularization. VIBNet~\cite{dai2018compressing} utilizes the variational information bottleneck principle to compress neural networks. They compel the networks to minimize the neuron redundancy across adjacent layers with binary mask vectors.
Beta-Bernoulli Dropout (BBDropout)~\cite{lee2018adaptive} learns a structured dropout function sampled from the Bernoulli distribution where the probability is given from the beta distribution with learnable parameters. Further, they introduce a data dependent BBDropout which generates a pruning mask as a function of the given data instance.

\paragraph{Meta learning.} Meta-learning, which learns over a distribution of task, have shown its efficiency in handling unseen tasks for various tasks, such as few-shot learning and sample-efficient reinforcement learning. The most popular meta-learning methods are gradient-based approaches such as MAML~\cite{finn2017model} and REPTILE~\cite{nichol2018reptile}, which aim to find an initialization that can rapidly adapt to new tasks. 
BASE~\cite{shaw2019meta} learns through MAML algorithm to rapidly search for optimal neural architecture, and thus significantly reduce the search cost over state-of-art neural architecture search (NAS) methods~\cite{liu2018darts,xie2018snas}. BASE learns a general prior through meta learning to perform fast adaptation for unseen tasks. On the other hand, our method learns the good initialization as a function of a set, such that it can rapidly \emph{adapt} to the given targe task. MetaPruning~\cite{liu2019metapruning} trains a hypernetwork that can generate sparse weights for any possible structures(i.e. the number of channels) of a network architecture. However, the hypernetwork does not generalize across tasks and thus the method requires to train one hypernetwork for each task.

\section{Rapid Structural Pruning of Neural Networks with Set-based Task-Adaptive Meta-Pruning}

We introduce a novel structural pruning method for deep neural networks, \emph{Set-based Task-Adaptive Meta-Pruning} (STAMP), which rapidly searches and prunes uninformative units/filters of the initial neural network trained on some other reference datasets. In Section~\ref{subsec:problem}, we define an optimization problem for deep neural networks with pruning masks. In Section~\ref{subsec:rapid-pruning}, we describe our set-based structural pruning method which efficiently reduces the model size in a few gradient steps while avoiding accuracy degradation. Finally, in Section~\ref{subsec:meta-update}, we describe our full meta-learning framework to train the pruning mask generator that generalizes to unseen tasks.

\subsection{Problem Definition}
\label{subsec:problem}
Suppose that we have a neural network $h(\mathcal{D}; \textbf{W})$, which is a function of the dataset $\mathcal{D}=\{\textbf{x}_i, \textbf{y}_i\}^N_{i=1}$ parameterized by a set of model weights $\textbf{W}=\{\textbf{W}_l\}^L_{l=1}$, where $\textbf{x}_i\in\mathbb{R}^{N\times X_d}$ and $l$ is a layer. Further suppose that the network has maximum desired cost $\kappa$ (e.g., FLOPs, Memory, the number of parameters, and training/inference time), which depends on the hardware capability and applications. By denoting the total cost of the model as $\mathcal{C}$, we formulate the problem of searching for a network that minimizes the task loss while satisfying the 
total cost $\mathcal{C}$ as an optimization problem, as follows:
\begin{align}
\begin{split}
\underset{\textbf{W}}{\mbox{minimize}}\frac{1}{N}\sum^N_{i=1}\mathcal{L}((\textbf{x}_i,\textbf{y}_i);\textbf{W})+\mathcal{R}(\textbf{W}),~~~\mbox{s.t.}~~|\mathcal{C}|\leq\kappa
\label{eq:cross-entropy}
\end{split}
\end{align}
where $\mathcal{R}$ is an arbitrary regularization term. To obtain an optimal model with the desired cost, we basically follow popular pruning strategy that adopts sparsity-inducing masking parameters for deep neural networks. We reformulate the problem as obtaining compressed weights $\bsy\omega_l$ with the corresponding binary masks $\textbf{m}_l$ at layer $l$, $\bsy\omega_l=\textbf{m}_l\odot\textbf{W}_l$, where $\textbf{m}_l\in\{0,1\}^{I_l\times C_l}$. This will result in \emph{unstructured pruning}, which will prune individual weight elements. However, we may allow the model to compress its size by \emph{structured pruning}, to yield actual wall-clock time speedup in training/inference time. We focus on generating structural pruning masks where the compressed weights will be expressed as  $\bsy\omega_l=\textbf{m}_l\otimes\textbf{W}_l$, where $\textbf{m}_l\in\{0,1\}^{C_l}$. Then, the objective function is defined to minimize a following loss function: $\mathcal{L}(\mathcal{D};\bsy\omega)+\mathcal{R}(\bsy\omega)$ where $\bsy\omega=\{\bsy\omega_l\}^L_{l=1}$.

\subsection{Rapid Structural Pruning with Set-encoded Representation}
\label{subsec:rapid-pruning}
\begin{figure}
\small
    \centering
        \includegraphics[height=2.5cm]{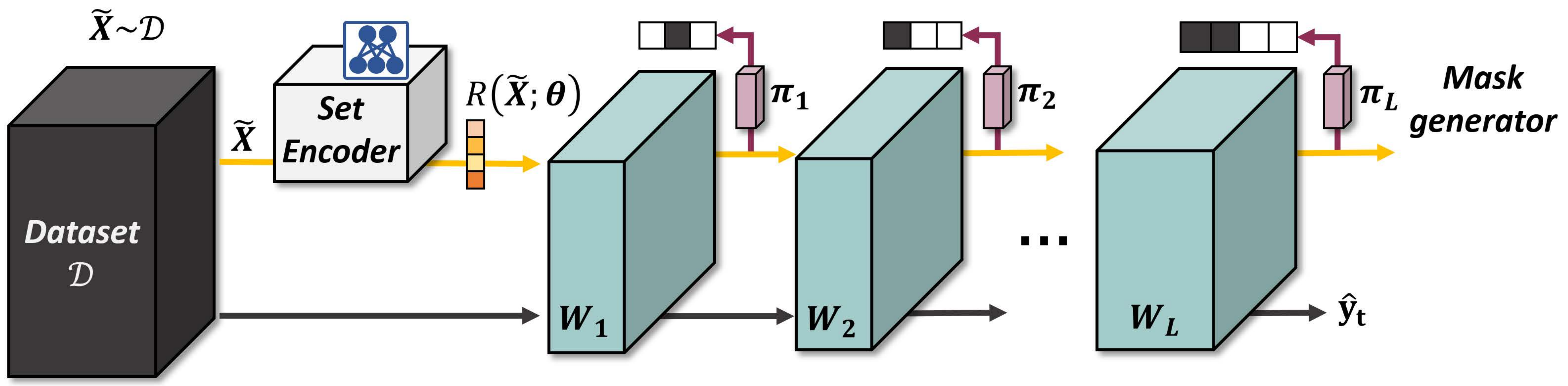}\\
    \vspace{-0.1in}
    \caption{\small \textbf{Set-based Task-Adaptive Pruning:} We sample the subset $\tilde{X}$ from $\mathcal{D}$ and train the model while simultaneously optimizing set-based binary masks through a set encoding function and mask generative functions.}
    \label{fig:method}
    \vspace{-0.1in}
\end{figure}

To obtain an optimal pruned structure for the target task, we need to exploit the knowledge of the given task. Conventional pruning schemes search for the desired subnetworks through full mini-batch training where all of the instances are trained through numerous iterations, incurring excessive training cost $\mathcal{C}$ as the data size gets bigger. To bypass this time-consuming search, and rapidly obtain the task-adaptive pruning masks, we adopt two learnable functions: a \textbf{set encoding function} $e(\mathcal{D};\bsy\theta)$ generates a set encoded output and a \textbf{mask generative function} $g(\cdot;\bsy\pi)$ obtains a binary mask vector $\textbf{m}$, parameterized by $\bsy\theta$ and $\bsy\pi$, respectively. That is, at each layer $l$, through two different functions, the model generates the task-adaptive mask vector $\textbf{m}_l$ given the dataset-level encoded representation from a set encoding function. To reduce an burden for encoding the entire dataset, we use a sampled subset $\tilde{\textbf{X}}\in\mathbb{R}^{B\times X_d}\sim\{\textbf{x}_i\}^N_{i=1}$ from $\mathcal{D}$, where $B<N$ is the sampled batch and $X_d$ is the input dimensionality.
To this end, we formulate the objective of our set-based task-adaptive pruning as follows:
\vspace{-0.05in}
\begin{align}
\begin{split}
\underset{\textbf{W},\bsy\theta,\bsy\pi}{\mbox{minimize}}\frac{1}{N}\sum^N_{i=1}\mathcal{L}((\textbf{x}_i,\textbf{y}_i);\bsy\omega)+\mathcal{R}(\bsy\omega),~~~|\mathcal{C}|\leq\kappa,
\label{eq:set-based-pruning}
\end{split}
\end{align}
\vspace{-0.15in}
\begin{align}
\begin{split}
\bsy\omega_l=g_l(\textbf{o}_l;\bsy\pi_l)\otimes\textbf{W}_l,~~\textbf{o}_l=h_l(\textbf{o}_{l-1};\textbf{W}_l),~~~\mbox{s.t.}~ ~\textbf{o}_0=e(\tilde{\textbf{X}};\bsy\theta)\in\mathbb{R}^{r\times X_d}
\label{eq:mask-generation}
\end{split}
\end{align}
where $r$ is a batch dimension of the set representation. Throughout the paper, we use $r=1$. The illustration of the set-based task-adaptive pruning model is described in Figure~\ref{fig:method}.

\subsection{Meta-update for Unseen Target Tasks}
\label{subsec:meta-update}
Now we describe how we learn the optimal parameters $\bsy\theta$ and $\bsy\pi$ for set based pruning. The simplest approach we can use is performing gradient descent through back propagation, as our model is end-to-end differentiable. However, this only results in optimized parameters for a specific task $\mathcal{D}_S$, which will not allow us to obtain an optimized parameters for an unseen task. As stated earlier, we want to apply our method across different task domains, such that we learn the pruning mask generator and the set encoder on $\mathcal{D}_S$ and apply them on $\mathcal{D}_T$. To this end, we apply gradient-based meta learning method which obtains initialization parameters that rapidly adapts to given (unseen) tasks. 

Basically, we train the parameters on multiple tasks sampled from $\mathcal{D}_S$ 
by computing inner gradient in the inner loop and combining them to update the outer loop parameters. Then, the objective of the meta-train step of \textit{STAMP} is learning good initialization of $\bsy\phi$ in the outer loop. We sample $N$ tasks from $\mathcal{D}_S$, where each task is $\mathcal{D}_S^{(n)}\subset\mathcal{D}_S$. From each $N$ tasks, a batch~$\{\textbf{x}^{(n)}_{j}, \textbf{y}^{(n)}_{j}\}^B_{j=1}$ is sampled and it is divided into mini-batches to update inner gradient in respect to the loss function $\mathcal{L}$ and the regularization terms described in Section~\ref{subsec:rapid-pruning}. We note that the whole batch excluding labels $\tilde{\textbf{X}}^{(n)}=\{\textbf{x}^{(n)}_{j}\}_{j=1}^B$ is used for encoding set representation. For updating outer loop parameters $\bsy\phi_i=\{\textbf{W}_i, \bsy\pi_i, \bsy\theta_i\}$ at $i^{th}$ epoch, we only use the  gradients of the last mini-batch, similarly to first-order MAML~\cite{finn2017model} to accelerate learning as below:
\begin{align}
\begin{split}
\bsy\phi_{i} \leftarrow \bsy\phi_{i-1} - \eta \sum_{n=1}^N \nabla_{\bsy\phi^{(n)}_i}L(\mathcal{D}_S^{(n)};\bsy\phi^{(n)}_i)
\end{split}
\end{align}
After meta learning a set of the parameters, we can adapt it to various unseen tasks $\mathcal{D}_T$ by performing few steps of gradient updates, with the maximum steps of 1 epoch. Through the meta-learn procedure, we can speed-up the training time on the target task by starting with the pruned network architecture in the early stage. We describe our whole process in Algorithm~\ref{alg:main_alg}. 

While we can plug in various set encoding methods ~\cite{edwards2016towards,zaheer2017deep} or pruning methods to the proposed framework, STAMP adopts a transformer module~\cite{lee2018set} for set encoding function $e(\cdot)$ and proposed a set-based pruning mask generator $g(\cdot)$ based on the Beta-Bernoulli dropout~\cite{lee2018adaptive}. The details of the set encoder and the structural mask generation function are described in the Appendix (Section~\ref{sub:app1}).

\begin{algorithm}[t]
\small
\caption{\modeltitle~(\shorttitle)}
\label{alg:main_alg}
\begin{algorithmic}[1]
\INPUT \mbox{Source Dataset}~$\mathcal{D}_S$, \mbox{Target Dataset}~$\mathcal{D}_T$,~\mbox{Learnable parameters}~~$\bsy\phi_{0}=\{\textbf{W}_{0},~\bsy\pi_{0},~\bsy\theta_{0}\}$
\FUNCTION{STAMP}
\FOR{$i=1,2,...$}
	\FOR{$n=1,...,N$ $\textbf{in parallel}$}
	    \STATE Sample task $\mathcal{D}_{S}^{(n)}\subset\mathcal{D}_{S}$, and $\bsy\phi_{i}^{(n) }=\bsy\phi_{i-1}$
    	\STATE Sample batch of tasks $\tilde{\textbf{X}}^{(n)}= \{\textbf{x}^{(n)}_{j}\}_{j=1}^B$,~~$\{\textbf{x}^{(n)}_{j}, \textbf{y}^{(n)}_{j}\}^B_{j=1}\sim\mathcal{D}^{(n)}_{S}$
	
		\STATE Compute $\mathcal{L}(\mathcal{D}_S^{(n)};\bsy\phi^{(n)}_i))$ with Eq.~\ref{eq:set-based-pruning}
			
		\ENDFOR
\STATE Update $\bsy\phi_{i} \leftarrow \bsy\phi_{i-1} - \eta \sum_{n=1}^N \nabla_{\bsy\phi^{(n)}_i}L(\mathcal{D}_S^{(n)};\bsy\phi^{(n)}_i))$
\ENDFOR
\ENDFUNCTION
\STATE  Meta train $\bsy\phi$  with function STAMP
\STATE  Prune $K$ step to optimize
$ \mathcal{L}(\mathcal{D}_T; \bsy\phi) $ with Eq.~\ref{eq:set-based-pruning}
\STATE  Finetune the pruned architecture to minimize 
$\mathcal{L}(\mathcal{D}_T;\bsy\omega)$
\end{algorithmic}
\end{algorithm}

\section{Experiments}
We demonstrate the effectiveness of STAMP with two widely used network architectures, namely VGGNet-19~\cite{zhuang2018discrimination} and ResNet-18~\cite{he2016deep}, on two benchmark datasets (CIFAR-10 and SVHN). We implement the code for all the experiments in Pytorch framework, and use Titan XP GPU for measuring the wall-clock time.

\paragraph{Baselines.} We validate our STAMP against recent structured pruning methods as well as unstructured random weight pruning methods. We also report the results on the variant of STAMP that only searches for the structure and randomly reinitializes the weights (STAMP-Structure). Baselines we use for comparative study are as follows:
\textbf{1) MetaPruning}~\cite{liu2019metapruning}: Structured pruning method which learns hypernetworks to generate pruned weights at each layer, and searches for the optimal pruned structure using an evolutionary algorithm. \textbf{2) BBDropout}~\cite{lee2018adaptive}: Beta-Bernoulli Dropout which performs structured pruning of the channels by sampling sparse masks on them. 
\textbf{3) Random Pruning}: Randomly pruning of channels. We sample the random structure ( i.e. the number of channels for each layer ) under the given FLOP constraints in the same manner as in MetaPruning~\cite{liu2019metapruning}. 
\textbf{4) Edge-Popup}~\cite{ramanujan2019s}: Unstructured pruning method that searches for the best performing sub-network of a network with random weights.
\textbf{5) SNIP}~\cite{lee2018snip}: One-shot unstructured pruning on random weights. We also report the results on a variant of SNIP which starts from pretrained weights (SNIP (P)). 
 For finetuning, we follow the standard setting from Zhuang et al.~\cite{zhuang2018discrimination} and perform mini-batch SGD training for 200 epochs where the batch size is set to 128. 

\paragraph{Networks and datasets}
As for the base networks, we use a modified version of \textbf{VGGNet-19} with 16 convolution layers and a single fully connected layer, and \textbf{ResNet-18} with an additional $1\times1$ convolution layer on the shortcut operation to resolve the dimensionality difference between pruned units/filters. We use VGG-19 and ResNet-18 trained on CIFAR-100 as the global reference network, and use CIFAR-10 and SVHN as the target tasks for evaluation of the pruning performance. 
 
 \paragraph{Meta-training}
 We meta-train our pruning mask generator on CIFAR-100 dataset. During meta-training time, we divide CIFAR-100 into 10 tasks (subsets), each of which contains 10 disjoint classes, and sampled 64 instances per class. We used total of 640 instances as the input to the set function to generate a set representation for each task. We also used the sampled instances for model training, by dividing it into 5 batches (128 instances for each). We used first-order MAML with Adam optimizer for both inner and outer parameter updates. 
 
 For more details on training of the baseline methods and meta-training for STAMP, such as learning rate scheduling, please see the Appendix (Section~\ref{sub:app2}).
 
 
\subsection{Quantitative Evaluation}
\label{subsec:quantative}
We report the results of pruning VGGNet-19 on CIFAR-10 and SVHN in Table~\ref{tab:vgg}, and ResNet-18 on CIFAR-10 in Table~\ref{tab:res}. We compare the accuracy as well as wall-clock training and inference time for all models at similar compression rate (Parameter Used or FLOPs). 

\begin{table*}
    \tiny
    
\begin{center}
\caption{\small Experiment results of CIFAR-10 and SVHN on VGGNet. \textbf{Training Time} consists of time to search for pruned network, and finetuning (200 epochs). \textbf{Expense} is computed by multiplying the training time by 1.46~\$, which is cost of using GPU (Tesla P100) on Google Cloud. The methods are sub-divided into the full network without pruning, unstructured pruning methods, structured pruning methods and \shorttitle.~\textbf{P} is a remaining parameter ratio. We run each experiments 3 times and report the mean $\pm$ std values.}

\small
\begin{tabular}{ c|| l |c|c|c|c|c|c }

\hline 
\multicolumn{1}{c||} {Datasets}&\multicolumn{1}{l|}{Methods}&\multicolumn{1}{c|}{Accuracy (\%)}&\multicolumn{1}{c|}{P (\%)}&\multicolumn{1}{c|}{FLOPs}&\parbox[c]{1.2cm}{ Training\\Time}&\parbox[c]{1.2cm}{Inference\\Time}&\multicolumn{1}{c}{Expense} \\
\hline \hline
& Full Network  & 93.72 $\pm$ 0.07 & 100 & x1.00 & 0.78 h  &0.85 sec &1.13 \$\\ \cline{2-8} 


&SNIP (P)~\cite{lee2018snip} & 92.98 $\pm$ 0.22  & 4.17 &x1.00 & 0.83 h &  0.92 sec &1.21 \$\\
&SNIP~\cite{lee2018snip}  & 92.85 $\pm$ 0.24 &4.17 	&x1.00 &0.83 h &  0.92 sec   &1.21 \$\\

\cdashline{2-8}

&Random Pruning & 92.01 $\pm$ 0.29  &32.20 	&x3.33 & 0.43 h  & 0.42 sec &0.62 \$\\
CIFAR-10& MetaPruning~\cite{liu2019metapruning} &92.12 $\pm$ 0.47	&21.84  &  x3.58	&4.99 h  &0.41  sec&7.28 \$\\

&BBDropout~\cite{lee2018adaptive}& 92.97 $\pm$ 0.10	&3.99  &  x3.42	&2.07 h & 0.43  sec&3.02 \$\\

\cdashline{2-8}

&STAMP-Structure&92.69 $\pm$ 0.13 & 4.43 & x3.48 &\textbf{0.44 h}&\textbf{0.36  sec}&\textbf{0.64 \$}\\
&STAMP&\textbf{93.49 $\pm$ 0.04} & 4.16 & x3.56 &\textbf{0.44 h}&  \textbf{0.36  sec} &\textbf{0.64 \$}\\


\hline
\hline

&Full Network			&
  95.99 $\pm$ 0.07& 100 & x1.00 &1.21 h &2.42  sec &1.76 \$\\
\cline{2-8}

&SNIP (P)~\cite{lee2018snip}	
& 95.56 $\pm$ 0.09  & 3.08 & x1.00& 1.22 h &2.45  sec&1.78 \$\\
&SNIP~\cite{lee2018snip} & 
95.52 $\pm$ 0.10 & 3.08 & x1.00 & 1.22 h &2.45  sec  &1.78 \$\\
\cdashline{2-8}

&Random Pruning& 95.56 $\pm$ 0.12& 28.95 &x3.40& 0.62 h &  1.27  sec &0.90 \$\\


SVHN

&MetaPruning~\cite{liu2019metapruning}& 95.50 $\pm$ 0.07 & 22.04 &  x3.64	&2.08 h & 1.44  sec&3.03 \$\\

& BBDropout~\cite{lee2018adaptive}	&\textbf{95.98 $\pm$ 0.19 }& 2.15 &x9.67&3.05 h &0.86  sec&4.45 \$\\

\cdashline{2-8}
&STAMP-Structure&95.39 $\pm$ 0.15& 3.08 &x4.60&\textbf{0.58 h}&0.91  sec&\textbf{0.84 \$}\\
&STAMP&\textbf{95.82 $\pm$ 0.16 }& 2.87 &x5.10& \textbf{0.58 h}&0.91  sec&\textbf{0.84 \$}\\



\end{tabular}
\label{tab:vgg}
\vspace{-0.2in}
\end{center}
\end{table*}



\begin{figure*}
\small
    \centering
    \begin{tabular}{c c c}\hspace{-0.1in}
        \includegraphics[height=3.3cm]{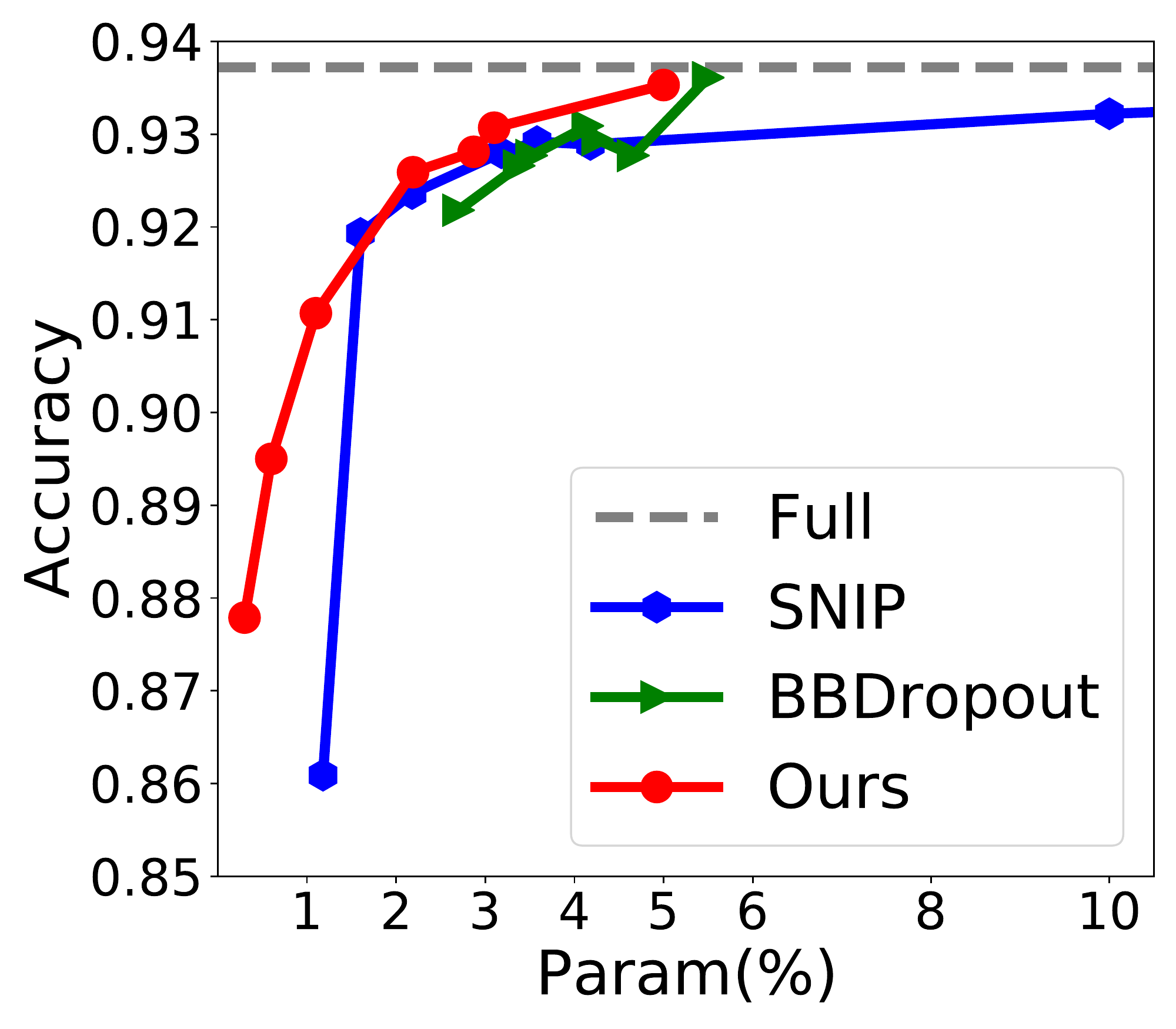}&
        \includegraphics[height=3.3cm]{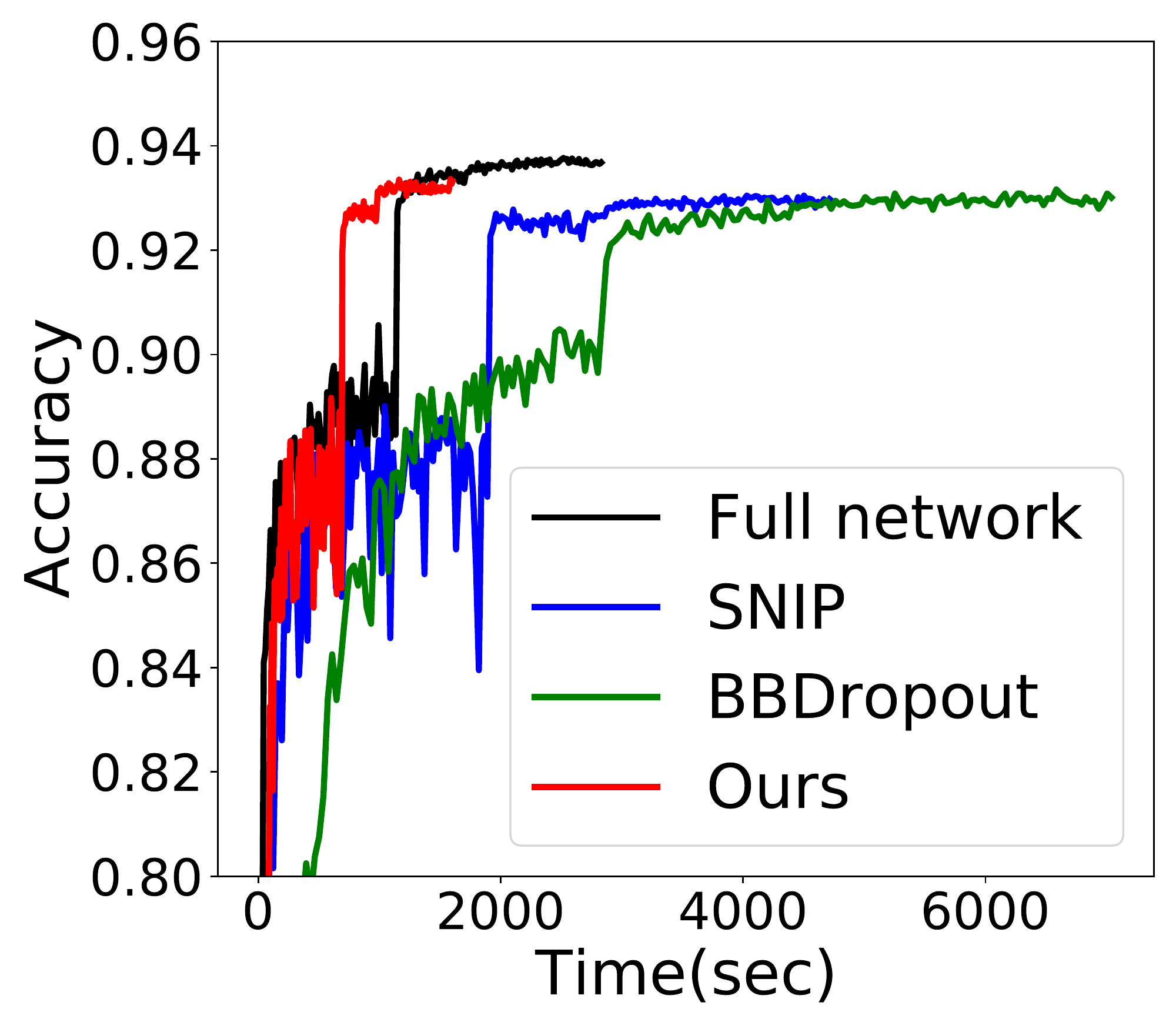}&
        \includegraphics[height=3.3cm]{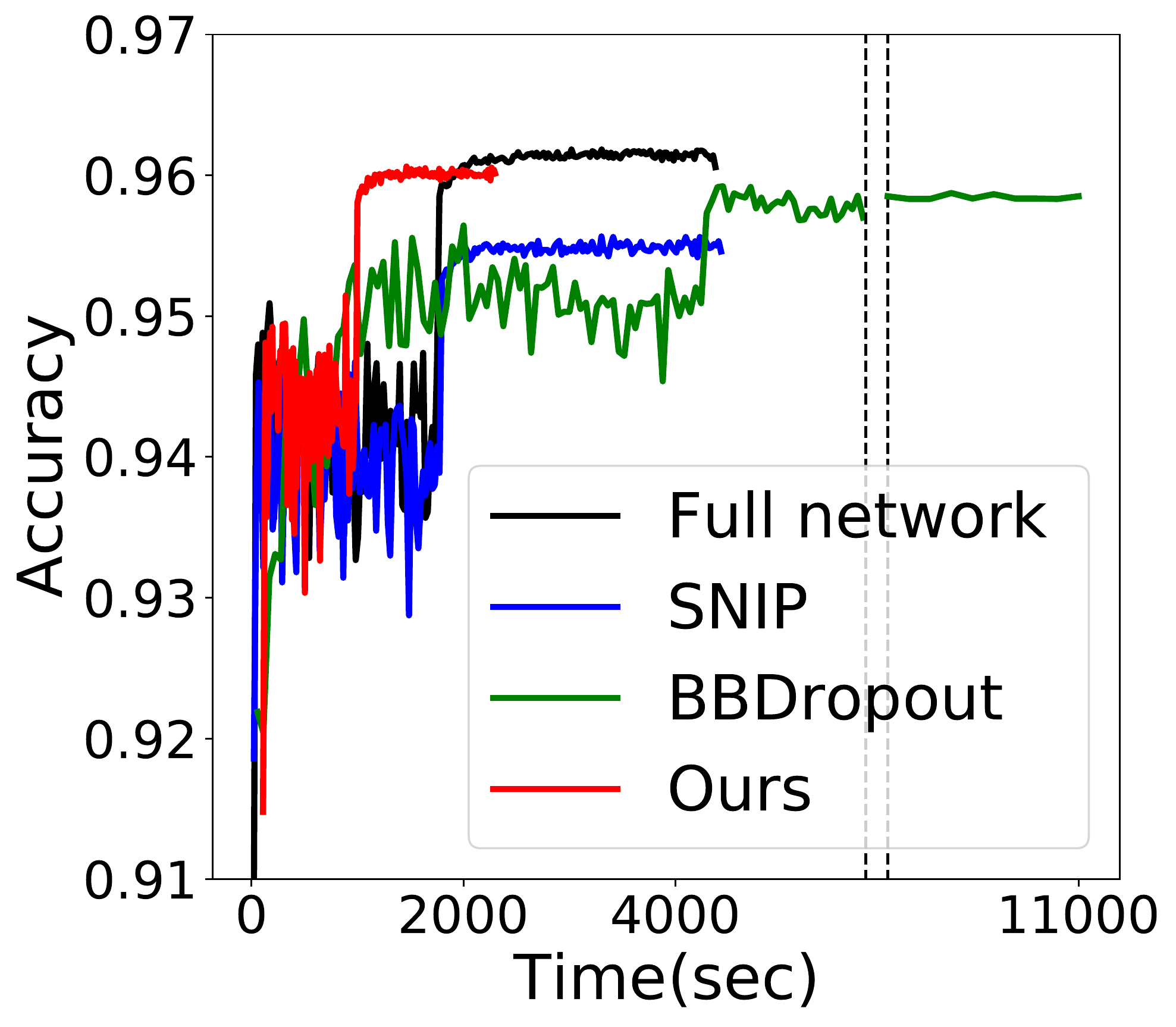}\\
        (a) Acc. over sparsity (CIFAR-10)&  (b) Acc. over time (CIFAR-10) &(c) Acc. over time (SVHN)  \\
    \end{tabular}
    \caption{\small \textbf{(a):} \textbf{Accuracy over the ratio of used parameters} for CIFAR-10 on VGGNet. \textit{Full} denotes the accuracy of the VGGNet before pruning. \textbf{(b-c):} \textbf{Accuracy over training time} for CIFAR-10 and SVHN.}
    \label{fig:fast}
    \vspace{-0.1in}
\end{figure*}

\paragraph{Accuracy over memory efficiency and FLOPs.} 
We first compare the accuracy over the parameter usage and theoretical computation cost, FLOPs. In Table~\ref{tab:vgg} and Table~\ref{tab:res}, SNIP with either random networks (SNIP) or the pretrained reference network (SNIP(P)) significantly reduce the number of activated parameters with a marginal drop of the accuracy for both CIFAR-10 and SVHN dataset. However, as the methods perform unstructured pruning, they can not reduce FLOPs which remains equal to the original full networks. On the other hand, structural pruning approaches show actual FLOPs reduction by pruning a group of weights (e.g. units/filters). Interestingly, MetaPruning, which applies a learned hypernetwork on a reference architecture and dataset to prune for the target dataset, obtains suboptimal architectures which sometimes even underperforms randomly pruned networks. This shows that the learned hypernetwork does not generalize across task domains, which is expected since it is not trained with diverse tasks. BBDropout achieves superior performance over other baselines with high model compression rate, but it requires large amount of training time to train the pruning mask generator, and thus slows down the training process over training of the full network. On the other hand, our STAMP either outperforms or achieve comparable performance to all baselines, in terms of both accuracy and compression rate. We further report the accuracy-sparsity trade-off for SNIP, BBDropout, and STAMP (Ours) in Figure~\ref{fig:fast} (a). Our method achieves better accuracy over similar compression rates, and shows marginal performance degeneration even with $1\%$ of the parameters remaining. Such good performance on unseen dataset is made possible by meta-learning the pruning mask generator.

\paragraph{Accuracy over wall-clock time for training/inference.} 
As described earlier, our main focus in this work is to significantly reduce the training time by obtaining a near-optimal compact deep networks for unseen targets on the fly, which is not possible with any of the existing approaches. As shown in Table~\ref{tab:vgg} and Table~\ref{tab:res}, unstructured random weights pruning methods (SNIP and Edge-Popup) do not results in any speedups in training time, and sometimes increases the cost of training over the full networks (See CIFAR-10 results in Table~\ref{tab:vgg}). These results are consistent with the findings in Frankle and Carbin~\cite{frankle2018lottery}, which showed that most of the subnetworks require larget number of training iterations over the full network. 

While structured pruning methods yield speedups in inference time over the full networks, MetaPruning and BBDropout need $\times2.01-\times2.76$ and $\times1.72-\times5.96$ more training time than full networks to search pruned architectures, respectively. On the contrary, STAMP instantly obtains a good subnetwork (single or less than $10$ iterations according to the pruned ratio), which trains \emph{faster} than the full network. STAMP is remarkably efficient over other structural pruning baselines, achieving $\times3.5-\times10.43$ and $\times2.81-\times5.76$ speedups over MetaPruning and BBDropout, respectively, with better or comparable performance. We further report the accuracy over training time for SNIP, BBdropout, and STAMP (Ours) in Figure~\ref{fig:fast} (b) and (c). Since our philosophy is train-once, and use-everywhere, once the mask generator is meta-learned on a reference dataset, it can be applied to any number of tasks without additional cost. Thus we excluded the meta training time of \shorttitle (15h on VGGNet and 30h on ResNet) and MetaPruning (1.2h) per task in Table~\ref{tab:vgg} and Table~\ref{tab:res}. 



\begin{table*}
\tiny
\begin{center}
\caption{\small Experiment results of CIFAR-10 on ResNet18.  Details are same with the Table~\ref{tab:vgg}. } 
\small
\begin{tabular}{l|| c|c|c|c|c|c }

\hline 
\multicolumn{1}{l||}{Methods}&\multicolumn{1}{c|}{Accuracy (\%) }&\multicolumn{1}{c|}{P (\%)}&\multicolumn{1}{c|}{FLOPs}&\multicolumn{1}{c|}{Training Time }&\multicolumn{1}{c|}{Inference time}&\multicolumn{1}{c}{Expense} \\
\hline \hline

Full Network	&
94.37 $\pm$ 0.12 & 100 & x1.00 & 1.08 h &1.02 sec &1.57 \$\\
\hline
Edge-Popup~\cite{ramanujan2019s}	&	
89.50 $\pm$ 3.46 &10.00 & x1.00& 1.38h &2.50 sec &2.01 \$\\ 

SNIP (P)~\cite{lee2018snip}	&	
93.17 $\pm$ 0.00 &10.04 & x1.00& 1.71 h &1.90 sec &2.49 \$\\ 

SNIP~\cite{lee2018snip}	&	
93.11 $\pm$ 0.00 &10.04 & x1.00&1.71 h &1.90 sec &2.49 \$\\   

\hdashline

Random Pruned&
91.95 $\pm$ 0.65 & 69.77 &  x3.65&0.58 h &0.58 sec &0.84 \$\\

MetaPruning~\cite{liu2019metapruning} &
91.01 $\pm$ 0.91& 66.02 &  x4.09 & 3.80 h  &0.58 sec &5.54 \$\\

BBDropout~\cite{lee2018adaptive} & 93.47 $\pm$ 0.14& 5.94 &x4.11&2.17 h&\textbf{0.54 sec} &3.16 \$\\

\hdashline

STAMP-Structure &\textbf{93.63 $\pm$ 0.08}  &9.07 & x4.08&\textbf{0.57 h} &\textbf{0.54 sec} &\textbf{0.83 \$}\\
STAMP&\textbf{93.61 $\pm$ 0.27} & 9.22 &x4.29 &\textbf{0.57 h} &\textbf{0.54 sec}  &\textbf{0.83 \$}\\


\end{tabular}
\label{tab:res}
\vspace{-0.2in}
\end{center}

\end{table*}

\begin{figure}
\vspace{-0.03in}
\begin{minipage}{.4\textwidth}
\begin{tabular}{c c}
\hspace{-0.2in}
\includegraphics[height=3cm]{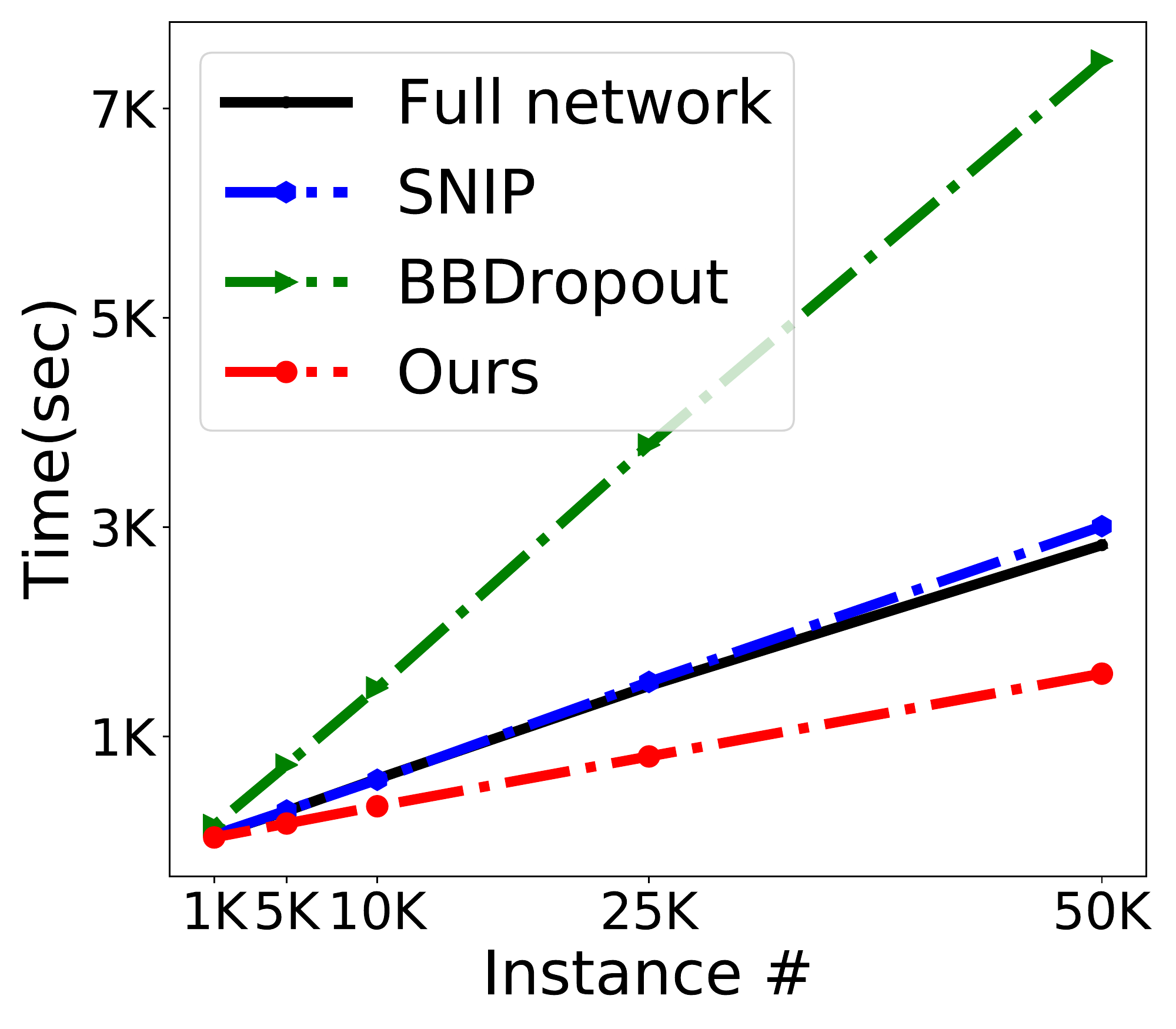} &
\hspace{-0.1in}\includegraphics[height=3.1cm]{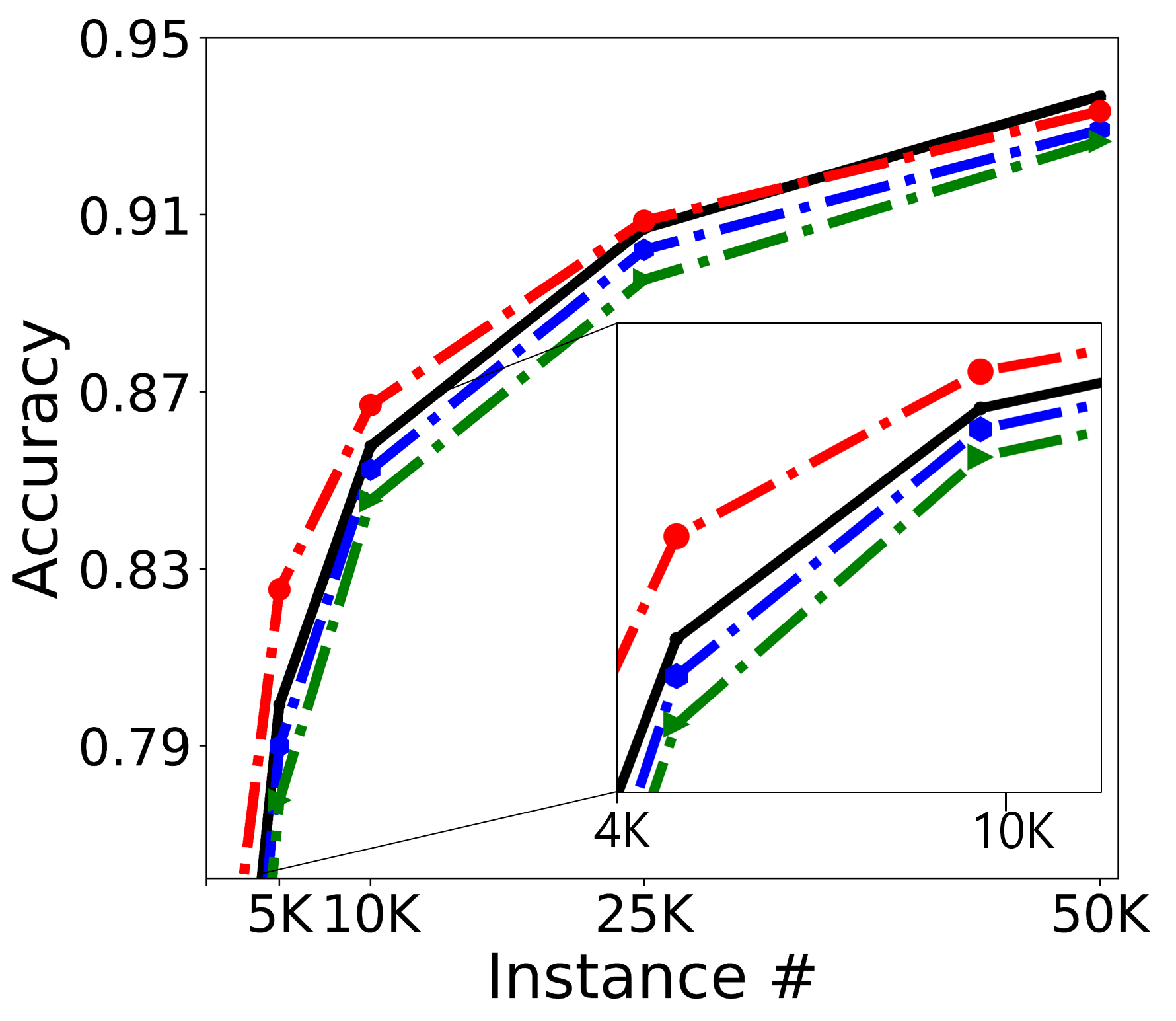}\\
\end{tabular}
\end{minipage}
\hspace{0.7in}
\begin{minipage}{.65\textwidth}
\footnotesize
\begin{tabular}{l||c|c|c|c}
\hline


\multicolumn{5}{c}{~~~~~~Accuracy (\%)}\\
\hline
\hline
 D Size & FULL & SNIP & BBD & STAMP \\
\hline
50K &
\textbf{93.68}&
92.92&
92.66&
93.34\\
25K &
90.69&
90.21&
89.53&
\textbf{90.86}\\
10K &
85.77&
85.24&
84.54&
\textbf{86.70}\\
5K &
79.93&
78.99&
77.77&
\textbf{82.53}\\
1K&
63.63&
60.34&
59.55&
\textbf{69.26}\\
\hline
\end{tabular}
\end{minipage}
\vspace{-0.1in}
\caption{\small{ \textbf{Left:} Training time over the number of training instances. \textbf{Middle:} Accuracy over the number of training instances, \textbf{Right:} Accuracy over the number of instances. All experimental results are obtained on CIFAR-10 with VGG-19. We compare our STAMP against Full Network~(\textbf{FULL}), SNIP~\cite{lee2018snip}, and BBDropout~\cite{lee2018adaptive}~(\textbf{BBD}). The sparsity is matched to prune $96 \% \sim 97\%$ of the parameters for \shorttitle~and SNIP.}}
\label{fig:n_clients}
\vspace{-0.2in}
\end{figure}


\paragraph{Data size of the target tasks.} We further examine the accuracy and time-efficiency of subnetworks obtained using different pruning methods on various problem size. We previously observed that STAMP can yield larger saving in the training and inference time as the network size gets larger (ReNet-18, Table~\ref{tab:res}). Another factor that defines the problem size is the number of instances in the unseen target dataset. We used subsets of CIFAR-10 to explore the effect of the task size to training time and accuracy in Figure~\ref{fig:n_clients}. The full dataset consists of 50K images, which corresponds to the results reported in Table~\ref{tab:vgg}. We observe that, as the number of instances used for training increases, STAMP obtains even larger saving in the training time, while BBDropout incurs increasingly larger time to train. Further, as the number of instances used for training becomes smaller, STAMP obtains larger gains in accuracy, even outperforming the full network, since the network will become relatively overparameterized as the number of training data becomes smaller. As another comparison with structural pruning method with learned masks, when using only 1K data instances for training, BBDropout finds the subnetwork attaining $43\%$ of parameters of the full network with $\times1.54$ FLOP speedup, while \shorttitle~prunes out $95.81\%$ of the parameters, resulting in $\times4.17$ speedup in FLOPs. This is because BBDropout learns the pruning mask on the given target task, and thus overfits when the number of training instances is small. STAMP, on the other hand, does not overfit since it mostly relies on the meta-knowledge and take only few gardient steps for the given task.

\subsection{Qualitative Analysis}
\vspace{-0.15in}
\paragraph{Pruned network structures.}
\begin{wrapfigure}{r}{3in}
\small
    \vspace{-0.1in}
    \begin{tabular}{cc}
        \hspace{-0.1in}\includegraphics[height=3.5cm]{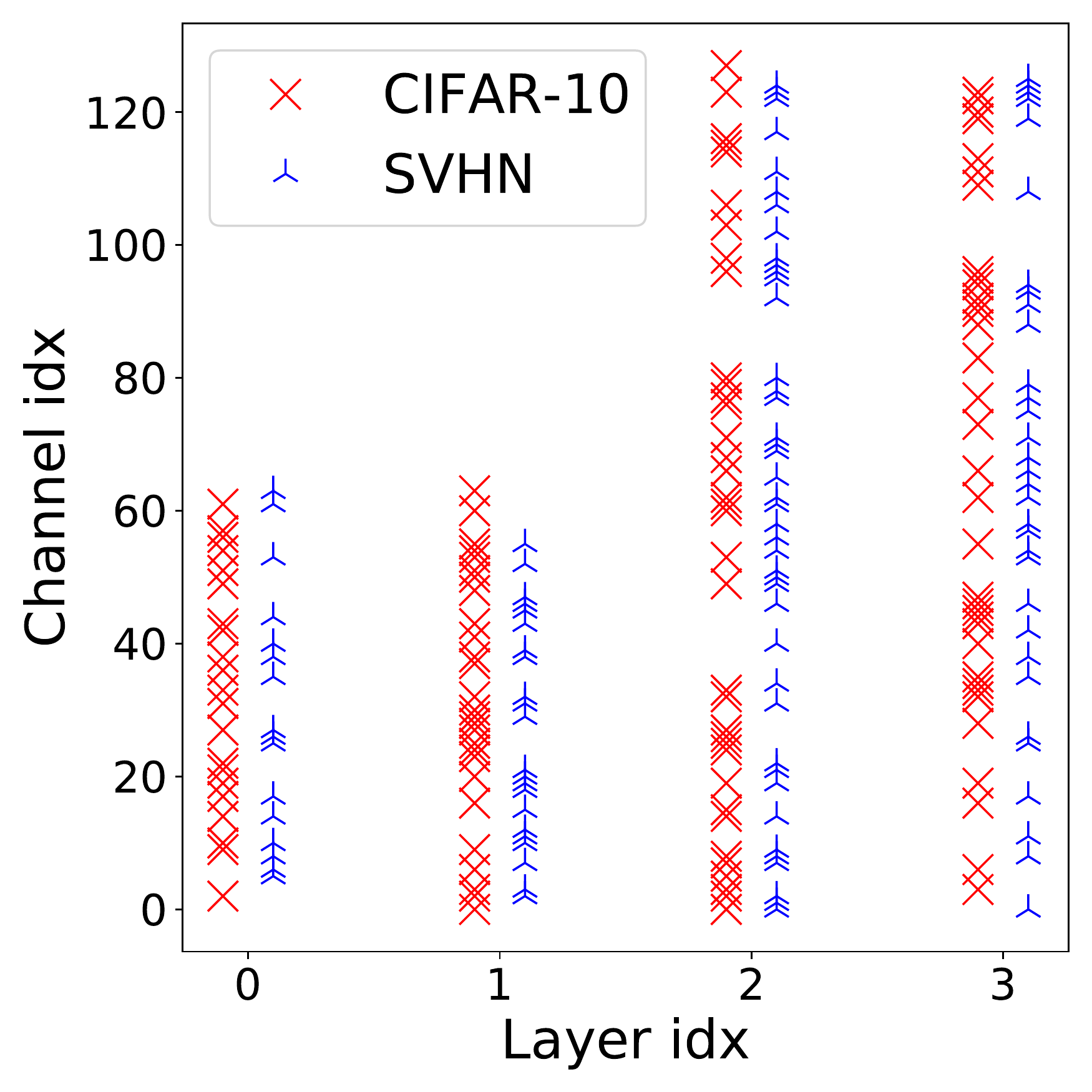} &\hspace{-0.1in}
        \includegraphics[height=3.5cm]{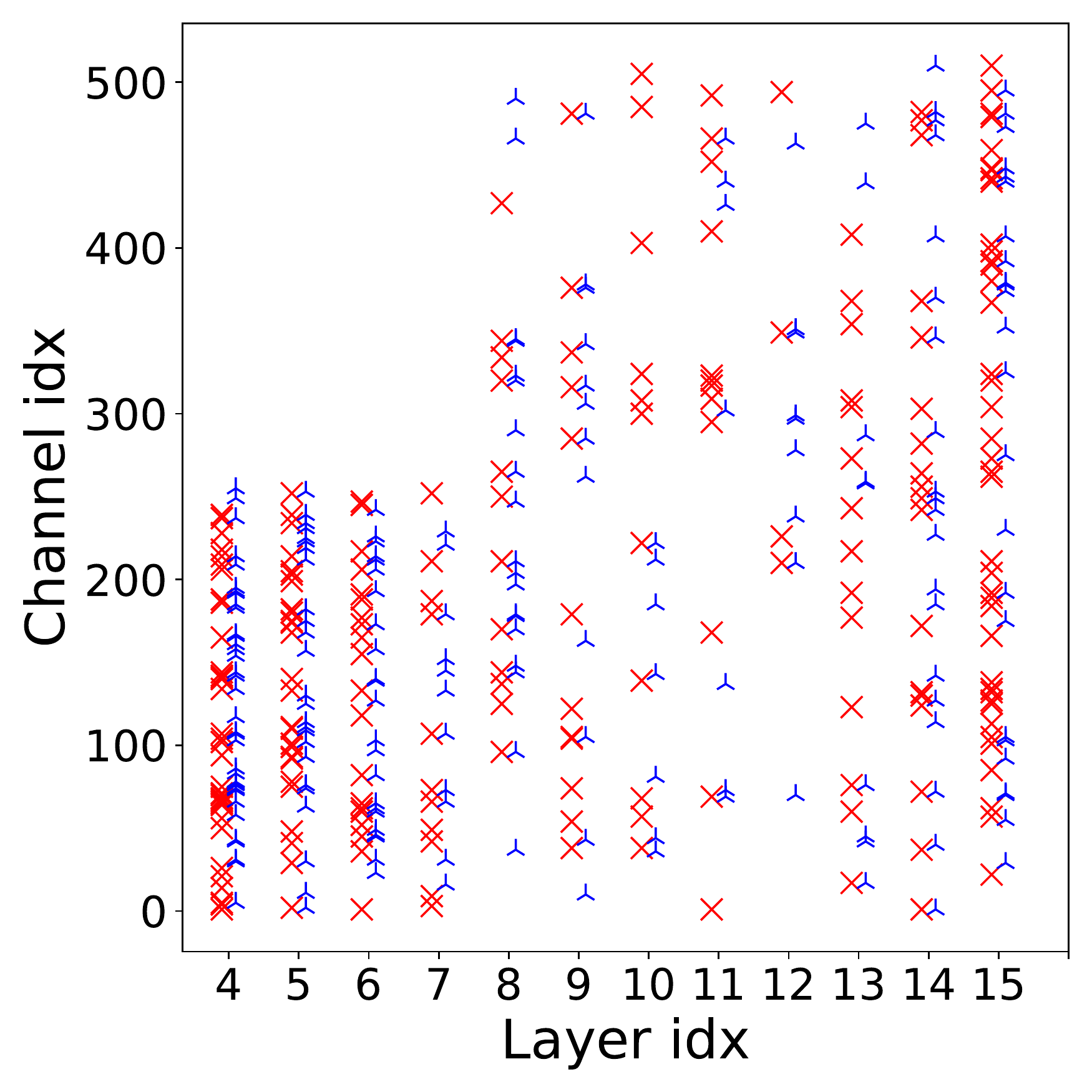}
        \\
        (a) Layer 0-3  &(b) Layer 4-15\\
    \end{tabular}
    \caption{\small We denote the indices of the remaining channels at each convolution layer of VGGNet after pruning on two different datasets, CIFAR-10 and SVHN. Both of them start from a same meta-learned parameters $\bsy\phi$.}
    \vspace{-0.15in}
    \label{fig:adaptive_pruning}
\end{wrapfigure} We further show the effect of task-adaptive pruning, which \shorttitle~will find different compressed subnetwork for different tasks. In other words, the pruning ratio and the pruned channels at each layer will be different for each dataset. We visualize the remaining channels of each convolution layer of VGGNet on CIFAR-10 and SVHN in Figure~\ref{fig:adaptive_pruning}. Note that unlike existing channel pruning methods, such as MetaPruning, we do not enforce any constraints (hyperparameters) on how much to prune, or what layer to prune since they are automatically determined by STAMP.

\section{Conclusion}
\vspace{-0.1in}
We proposed a novel set-based task-adaptive structural pruning method which instantly generates a near-optimal compact network for the given task, by performing rapid structural pruning of a global reference network trained on a large dataset. This is done by meta-learning a pruning mask generator over multiple subsets of the reference dataset as a function of a dataset, such that it can generate a pruning mask on the reference network for any unseen tasks. Our model, STAMP obtains a compact network that not only obtains good performance with large reduction in the memory and computation cost at inference time, but also enables \emph{training} time speedup which was not possible with previous methods. Further analysis showed that STAMP obtains larger performance gains when the target dataste is small, and prunes out the channels of the same reference network differently for each dataset. We believe that both the proposal of a framework that can obtain optimal compact network for unseen tasks, and achieving training time speedup are important contributions that enhances efficiency and practicality of pruning methods.

\section{Broader Impact}
Every day, a tremendous amount of computing resources are used for training deep neural networks, and searching for the optimal architecture for the given task either manually or by automatic search with neural architecture search (NAS). Our method can significantly reduce the time and energy spent for both architecture search and training. 

\begin{itemize}
\item \textbf{Significant reduction in the architecture search cost.} By instantly generating the optimal network architecture (which is a subnetwork of a reference network) for a given dataset, our method allows to greatly reduce the architecture search time for each individual task. 
\item \textbf{Significant reduction in the training cost.} Our method largely reduces the training time required to train the network for a given task, as the obtained subnetwork will lead to actual saving in memory, computation, and wall-clock time required for training.
\end{itemize}

Such reduction in both architecture search and training time will allow monetary savings and minimize energy consumption, making deep learning more affordable to service providers and end-users that cannot manage the large cost of running the model on cloud or GPU clusters. 

\bibliography{refs}
\bibliographystyle{unsrt}

\newpage
\appendix

\section{Appendix}

\paragraph{Organization.} The supplementary file is organized as follows: We first describe each component of our Set-based Task-Adaptive Meta-Pruning (STAMP) in detail, including the set encoding function and the set-based structural pruning method (mask generative function) in Section~\ref{sub:app1}. Then, in Section~\ref{sub:app2}, we provide the detailed experimental settings and additional results on SVHN using ResNet-18 as the backbone network.

\section{Structural Binary Mask Generation with a Set-encoded Representation}
\label{sub:app1}
We now describe how we obtain the set representation with $e(\cdot;\bsy\theta)$ and learn structural pruning masks with the set-based mask generative function $g(\cdot;\bsy\pi)$ introduced in Section~\ref{subsec:rapid-pruning}.
\subsection{Set Encoding Function}


To obtain an optimally pruned neural architecture for the target task, we need to exploit the knowledge of the given task. Conventional pruning schemes search for the desired subnetworks through full mini-batch training, which will incur excessive training cost $\mathcal{C}$ when the data size is large. On the other hand, we rapidly and precisely obtain the desired pruned structures given the encoded reprsentations for each dataset . This procedure of obtaining a set representation $e(\tilde{\textbf{X}};\bsy\theta)$  with the set encoder $e$ parameterized by $\bsy\theta_{(\cdot)}$ is given as follows:
\begin{align}
\begin{split}
e(\tilde{\textbf{X}};\bsy\theta) = f_{D}(f_{E}(\tilde{\textbf{X}}; \bsy\theta_{E}); \bsy\theta_{D})~~~s.t.~~~\tilde{\textbf{X}}\in\mathbb{R}^{B\times X_d} \sim \{\textbf{x}_i\}_{i=1}^N,~\textbf{o}_0=e(\tilde{\textbf{X}};\bsy\theta)\in\mathbb{R}^{r\times X_d}
\label{eq:set_func}
\end{split}
\end{align}
where $\tilde{\textbf{X}}\in\mathbb{R}^{B\times X_d}$ from $\mathcal{D}$ is a sampled task set, $B\leq N$ is the sampled batch, $X_d$ is the input dimensionality, and $r$ is the batch dimension of the set representation ($r\ll B$). We then define the set function as a stacked function of the encoder and the decoder where $f_{E}(\cdot;\bsy\theta_{E})\in\mathbb{R}^{B\times X_d}$ is an arbitrary encoder function parameterized by $\bsy\theta_{E}$ and $f_{D}(\cdot;\bsy\theta_{D})$ is a decoder function parameterized by $\bsy\theta_{D}$. The encoder encodes the sampled task set $\tilde{\textbf{X}}$ and the decoder regenerates the encoded vector to the dataset-level representation. Throughout the paper, we use $r=1$ as the dimension of the set representation. We adopt a transformer module \cite{lee2018set} for set encoding, which is a learnable pooling neural network module with an attention mechanism as shown below:
\begin{align}
\begin{split}
f_E(\cdot)=\mbox{AE}(\mbox{AE}(\cdot)),~~~f_D(\cdot)=\mbox{rF}(\mbox{AE}(\mbox{Pool}_r(\cdot))).
\label{eq:transformer0}
\end{split}
\end{align}
where rF is a fully connected layer, AE is an attention-based block~\cite{vaswani2017attention}, and $\mbox{Pool}_r$ is a Pooling by Multihead Attention with $r$ seed vectors~\cite{lee2018set}. AE is a permutation equivariant block that constructs upon Multi-head attention (MH), introduced in Transformer~\cite{vaswani2017attention}. In other words, AE encodes the set information and consists $f_E(\cdot)$, while $f_D(\cdot)$ also includes AE to model correlation between $r$ vectors after pooling. AE is defined as below:
\begin{align}
\begin{split}
    \mbox{AE}(\cdot) =\mbox{Norm}(H + \mbox{rF}(H))\\
\end{split}
\end{align} 
\begin{align}
\begin{split}
    \mbox{where}~ H= \mbox{Norm} (\cdot + \mbox{MH}(\cdot,\cdot,\cdot))\\
\end{split}
\end{align}
where Norm is layer normalization~\cite{ba2016layer}. The encoder encodes the given dataset using the above module, and the decoder aggregates the encoded vector. The full encoding-decoding process can be described as follows:
\begin{align}
\begin{split}
e(\tilde{\textbf{X}};\bsy\theta) = \mbox{rF}(\mbox{AE}(\mbox{Pool}_r(\mbox{AE}(\mbox{AE}(\tilde{\textbf{X}})))))\in \mathbb{R}^{r\times X_d} 
\end{split}
\end{align}
\begin{align}
\begin{split}
\mbox{where}~ \mbox{Pool}_r(\cdot) = \mbox{AE}(R,\mbox{rF}(\cdot))\in \mathbb{R}^{r\times X_d}
\end{split}
\end{align}
In here, pooling is done by applying multihead attention on a learnable vector $R\in \mathbb{R}^{r\times X_d}$. We set $r=1$ in the experiments to obtain a single set representation vector. By stacking these attention based permutation equivaraint functions, we can obtain the set representation $e(\tilde{\textbf{X}};\bsy\theta)\in \mathbb{R}^{1\times X_d}$ from the sampled task $\tilde{\textbf{X}}\in\mathbb{R}^{B\times X_d} $.





\subsection{Mask Generative Function}

We now describe the mask generation function $\textbf{m}_l=g_l(\textbf{o}_l;\bsy\pi_l)$ at layer $l$, from which we obtain the pruned model parameter $\bsy\omega_l=\textbf{m}_l\otimes \textbf{W}_l$. 
Similarly as in Lee et al. \cite{lee2018adaptive}, we use the following sparsity-inducing beta-Bernoulli prior to generate a binary pruning mask at each layer, which follows Bernoulli distribution,  \textit{Bernoulli}$(\textbf{m}_l;\bsy\pi_l)$, given the probability of parameterized beta distribution as follows: 
\begin{align}
\begin{split}
\textbf{m}_l|\bsy\pi_l\sim\prod^{C_l}_{c=1}\mbox{Bernoulli}(\mbox{m}_{c,l};\pi_{c,l}),~~~
\mbox{where}~~\bsy\pi_l\sim\prod^{C_l}_{c=1}\mbox{beta}(\pi_{c,l};\alpha_l,1)
\label{eq:bbdrop}
\end{split}
\end{align}
where $C_l$ is the number of channels in layer $l$. With a learnable parameter $\alpha_l$ for the beta distribution, the model learns the optimal binary masks from a randomly sampled value from the beta distribution, to determine which neurons/channels should be pruned. 
We extend this input-independent pruning method to sample binary masks based on the set representation of the target task. This \textit{set-dependent} pruning with STAMP is different from data-dependent BBDropout~\cite{lee2018adaptive} in that the former generates a mask per dataset while the altter generates a mask per instance, which makes it difficult to globally eliminate a channel. Furthermore, rather than searching for the compressed structure by training with mini-batch SGD at each iteration, we utilize a set representation to rapidly obtain a near-optimal subnetwork within a few gradient steps. With the set $e(\tilde{\textbf{X}};\bsy\theta)$ representation obtained from the given dataset $\textbf{X}$, we  calculate the activation $\textbf{o}_l=h_l(\textbf{o}_{l-1};\textbf{W}_l)$ for each layer $l$, where $h_l$ is the function of the layer $l$ (i.e. convolution) and $\textbf{o}_1=h_1(e(\tilde{\textbf{X}};\bsy\theta);\textbf{W}_1)$. We omit the layer notation $l$ for readability in the following equations. Then, we sample a structural pruning mask vector $\textbf{m}$ as follows:
\begin{align}
\begin{split}
\textbf{m}|\bsy\pi,\textbf{o}\sim\prod^{C}_{c=1}\mbox{Bernoulli}(\mbox{m}_c;\pi_c\psi(\gamma_c o_c+\beta_c,\epsilon)),~~~\mbox{s.t.}~~[o_1;...;o_C]=\mbox{Pool}(\textbf{o}_l),
\label{eq:set-bbdrop}
\end{split}
\end{align}
where $\gamma$ and $\beta$ are learnable scaling and shifting factors and Pool is the average pooling for $\textbf{o}$ which obtains a representative value for each channel. The clamping function is defined as $\psi(\cdot,\epsilon)=\mbox{min}(1-\epsilon, \mbox{max}(\epsilon, \cdot))$ with a small $\epsilon>0$. Using a clamping function, the network will retain only the meaningful channels. We employ variational inference to approximate sparsity inducing posterior  $p(\textbf{W},\bsy\pi,\bsy\theta,\bsy\beta|\textbf{X})$. The KL-divergence term for our set-based task-adaptive pruning is as follows:
\vspace{0.1in}
\begin{align}
\begin{split}
D_{KL}[q(\textbf{m},\bsy\pi,\bsy\theta|\textbf{X})\|p(\textbf{m},\bsy\pi,\bsy\theta)] + D_{KL}[q(\bsy\beta)\|p(\bsy\beta)],~~~s.t.~~\bsy\beta\sim\mathcal{N}(\textbf{0},b\textbf{I}),
\label{eq:set-bbdrop-kl}
\end{split}
\end{align}
where $b$ is a fixed value for a variance of the shifting factor $\bsy\beta$ to prevent $\bsy\beta$ from drifting away. The first term can be computed analytically to obtain a closed form solution~\cite{lee2018adaptive,nalisnick2016stick}. Also, we can easily compute the second term, $\mathcal{R(\bsy\omega)}$ in the objective function of \shorttitle~(Equation ~\ref{eq:set-based-pruning}) by updating it with gradient-based methods. 

We can further approximate the expectation for the prediction of given dataset $\mathcal{D}$ as follows: 
\begin{align}
\begin{split}
p(\textbf{y}|\textbf{x},\textbf{W},\bsy\theta)\approx p(\textbf{y}|f(\textbf{x};\mathbb{E}_q[\textbf{m}]\otimes\textbf{W},\bsy\theta)),~~~s.t.~~\mathbb{E}_q[\textbf{m}]=\mathbb{E}_q[\bsy\pi]\cdot\psi(\bsy\gamma\mbox{Pool}(\textbf{o})+\bsy\beta,\epsilon)
\label{eq:set-bbdrop-approx}
\end{split}
\end{align}

\section{Experiments}
\label{sub:app2}

\subsection{Experimental Settings}
 We first describe how we meta-train \shorttitle~and set the settings for the baselines , SNIP~\cite{lee2018snip} and MetaPruning~\cite{liu2019metapruning}, for the experiments in the main paper (VGGNet and ResNet-18 on two benchmark CIFAR-10 and SVHN).

For \shorttitle, in \textbf{function} STAMP in Algorithm 1, we update $\bsy\phi=\{\textbf{W},~\bsy\pi,~\bsy\theta\}$ with the learning rate 0.001, 0.01, and 0.001 with Adam optimizer, while decreasing the learning rate by 0.1 at 50\% and 80\% of the total epoch, following the settings of BBDropout~\cite{lee2018adaptive}. For  Algorithm 1, we select $i=2000, B=640$ for VGGNet and $i=3000, B=500$ for ResNet-18. We sampled the same number of instances per class. We further set $N=10$ and the size of the minibatch as $32$. When pruning with \shorttitle, we use the same learning rate as the one we use in the meta training stage for VGGNet. However, for ResNet-18, we set the learning rates as $~\bsy\pi=0.5$ to adjust the pruning rate.

For SNIP~\cite{lee2018snip}, in the ResNet-18 experiment, we do not prune the $1\times1$ convolution layer to match the settings for STAMP experiments. Additionally, we modify the learning rate to 0.01, since at the learning rate of $0.1$, SNIP (P) and SNIP obtained lower accuracies (88.51\% and 85.26\% respectively). For VGGNet, we prune the weights of 16 convolution layers. For SNIP (P) we load the pretrained weights on CIFAR-100 before pruning. 

For MetaPruning~\cite{liu2019metapruning}, we used the same settings for ResNet-18 and ResNet-50 experiments. For VGGNet, we prune filters of 16 convolution layers which is the same as \shorttitle. At the search phase, we search for the architecture under given FLOP constraints. We set the pruning ratio at each layer between 20 \% to 60 \%, to satisfy the FLOP constraints, which is 40 \% to 80 \% for the given setting. For the rest of the experimental settings, we followed the settings in~\cite{liu2019metapruning}.

\subsection{Experimental Results} We report the experimental results on SVHN with ResNet18 in Table~\ref{tab:TAB}, which was omitted from the main paper due to the page limit. We followed the settings of Liu et al.~\cite{liu2017learning} and trained on SVHN for 20 epochs. All other settings are kept the same as the experimental setting in the previous paragraph. The results show that \shorttitle~ has the best trade-off between the accuracy and the efficiency.

\begin{table*}
\tiny
\begin{center}
\caption{\small Experiment results of SVHN on ResNet-18. \textbf{Training Time} consists of time to search for the pruned network and finetuning (200 epochs). \textbf{Expense} is computed by multiplying the training time by 1.46~\$, which is the cost of using GPU (Tesla P100) on Google Cloud. The methods are sub-divided into the full network without pruning, unstructured pruning methods, structured pruning methods, and \shorttitle~(\shorttitle-Structure is a variation of \shorttitle, which re-initializes the pruned architecture). \textbf{P} is the remaining parameter ratio. We run each experiment 3 times and report the mean $\pm$ std values. } 
\small
\begin{tabular}{l|| c|c|c|c|c|c }

\hline 
\multicolumn{1}{l||}{Methods}&\multicolumn{1}{c|}{Accuracy (\%) }&\multicolumn{1}{c|}{P (\%)}&\multicolumn{1}{c|}{FLOPs}&\multicolumn{1}{c|}{Training Time }&\multicolumn{1}{c|}{Inference time}&\multicolumn{1}{c}{Expense} \\
\hline \hline

Full Network	&
94.57 $\pm$ 0.01 & 100 & x1.00 & 0.16 h & 3.30 sec & 0.24 \$\\
\hline
Edge-Popup~\cite{ramanujan2019s}	&	
92.61 $\pm$ 0.01 & 5.00 & x1.00& 0.20 h & 6.15 sec &0.29 \$\\ 

SNIP (P)~\cite{lee2018snip}	&	
95.38 $\pm$ 0.01 & 6.06 & x1.00& 0.35 h &6.64 sec &0.51 \$\\ 

SNIP~\cite{lee2018snip}	&	
94.88 $\pm$ 0.01 & 6.06 & x1.00& 0.35 h &6.64 sec &0.51 \$\\

\hdashline

Random Pruned&
 94.39 $\pm$ 0.23 & 72.17 &  x2.99&\textbf{0.08 h} &1.66 sec &\textbf{0.12 \$}\\

MetaPruning~\cite{liu2019metapruning} &
94.49 $\pm$ 0.19 & 70.99 &  x2.83 & 2.41 h  &1.68 sec &3.51 \$\\

BBDropout~\cite{lee2018adaptive} & 94.32 $\pm$ 0.02& 4.90 &x5.25& 0.31 h& 1.52 sec & 0.46 \$\\

\hdashline

STAMP-Structure &\textbf{ 95.17 $\pm$ 0.01}  & \textbf{4.81} &x\textbf{5.47} &\textbf{0.11 h} &\textbf{1.51 sec}  &\textbf{0.16 \$}\\
STAMP&\textbf{ 95.41 $\pm$ 0.01} & \textbf{4.81} &x\textbf{5.47} &\textbf{0.11 h} &\textbf{1.51 sec}  &\textbf{0.16 \$}\\

\end{tabular}
\label{tab:TAB}
\vspace{-0.2in}
\end{center}

\end{table*}

\begin{figure}
\vspace{0.1in}
\begin{minipage}{.4\textwidth}
\begin{tabular}{c }
\hspace{-0.2in}
\includegraphics[height=3.3cm]{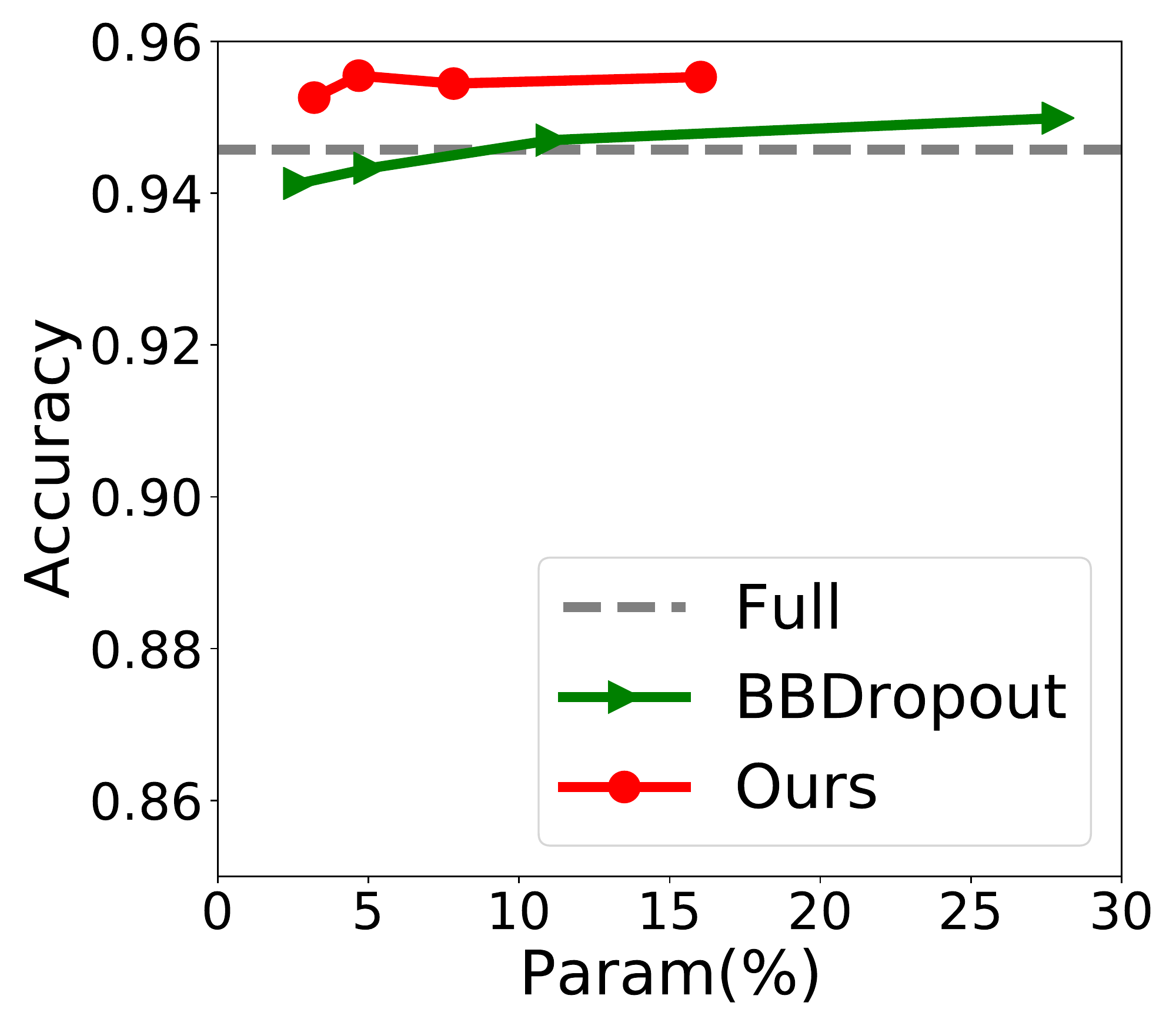}\\


\end{tabular}
\end{minipage}
\hspace{-0.7 in}
\begin{minipage}{.65\textwidth}
\footnotesize
\begin{tabular}{l|| c|c|c|c }

\hline 
\multicolumn{1}{l||}{Methods}&\multicolumn{1}{c|}{Accuracy (\%) }&\multicolumn{1}{c|}{P (\%)}&\multicolumn{1}{c|}{FLOPs}&\parbox[c]{1.5cm}{ Training\\Time}\\
\hline \hline

Full Network	&
94.57  & 100 & x1.00 & \textbf{0.16 h} \\
\hline

BBD ($kl$=15)& 94.70 & 11.09 &x2.86& 0.31 h\\

BBD ($kl$=20)& 94.30 & 4.86 &x5.14& 0.31 h\\
BBD ($kl$=25)& 94.12 & 2.71 &x8.62& 0.31 h\\

\hdashline

STAMP ($K$=1, $kl$=15)&\textbf{ 95.44 } & \textbf{7.83} &x\textbf{4.26} &\textbf{0.11 h} \\

STAMP ($K$=5)&\textbf{ 95.73 } & \textbf{3.84} &x\textbf{6.39} &\textbf{0.19 h} \\
STAMP ($K$=10)&\textbf{ 95.77 } & \textbf{1.81} &x\textbf{8.13} &\textbf{0.31 h} \\


\end{tabular}

\end{minipage}
\vspace{0.1in}
\caption{\small{ \textbf{Left:} Accuracy over the ratio of used parameters for SVHN on ResNet-18. \textit{Full} denotes the accuracy of the ResNet-18 before pruning. \textbf{Right:} Exploring different $K$ (the number of epochs of pruning stage) of ~\shorttitle~compared with BBDropout~\cite{lee2018adaptive}~(BBD). $kl$ is a scale factor for the regularization term. }}
\label{fig:FIG}
\vspace{-0.2in}
\end{figure}

STAMP obtains higher accuracy over BBDropout at the same compression rate as shown in Figure~\ref{fig:FIG}). Further, when trained for larger number of epochs, STAMP can obtain even higher accuracy and larger compression rate over BBD as shown in Figure~\ref{fig:FIG}, outperforming all baselines in Table~\ref{tab:TAB}. Although training STAMP for longer epochs yields slightly higher training time than the time required to train the full network, STAMP still trains faster than BBdropout.

\end{document}